\documentclass{article}

\usepackage{microtype}
\usepackage{graphicx}
\usepackage{subfigure}
\usepackage{booktabs} 
\usepackage{comment}
\usepackage{hyperref}

\usepackage[accepted]{icml2025}

\usepackage{amsmath}
\usepackage{amssymb}
\usepackage{mathtools}
\usepackage{amsthm}
\usepackage{subfigure}
\usepackage{multirow}
\usepackage{threeparttable, booktabs}
\usepackage{enumitem}
\usepackage{bbding}
\usepackage[capitalize,noabbrev]{cleveref}

\theoremstyle{plain}

\theoremstyle{definition}

\theoremstyle{remark}

\usepackage[textsize=tiny]{todonotes}

\icmltitlerunning{Expand Heterogeneous Learning Systems with Selective Multi-Source Knowledge Fusion}

\begin{document}

\twocolumn[
\icmltitle{Expand Heterogeneous Learning Systems with \\ Selective Multi-Source Knowledge Fusion}



\icmlsetsymbol{equal}{*}

\begin{icmlauthorlist}
\icmlauthor{Gaole Dai}{ntu}
\icmlauthor{Huatao Xu}{hkust}
\icmlauthor{Yifan Yang}{msra}
\icmlauthor{Rui Tan}{ntu}
\icmlauthor{Mo Li}{hkust}
\end{icmlauthorlist}

\icmlaffiliation{ntu}{College of Computing and Data Science, Nanyang Technological University, Singapore}
\icmlaffiliation{msra}{Microsoft Research, Shanghai, China}
\icmlaffiliation{hkust}{Hong Kong University of Science and Technology, Hong Kong}

\icmlcorrespondingauthor{Mo Li}{lim@ust.hk}
\icmlcorrespondingauthor{Rui Tan}{tanrui@ntu.edu.sg}

\icmlkeywords{Deep Learning Systems, Transfer Learning, Knowledge Distillation}

\vskip 0.3in
]



\printAffiliationsAndNotice{}  

\begin{abstract}
Expanding existing learning systems to provide high-quality customized models for more domains, such as new users, is challenged by the limited labeled data and the data and device heterogeneities. While knowledge distillation methods could overcome label scarcity and device heterogeneity, they assume the teachers are fully reliable and overlook the data heterogeneity, which prevents the direct adoption of existing models. To address this problem, this paper proposes a framework, HaT, to expand learning systems. It first selects multiple high-quality models from the system at a low cost and then fuses their knowledge by assigning sample-wise weights to their predictions. Later, the fused knowledge is selectively injected into the customized models based on the knowledge quality. Extensive experiments on different tasks, modalities, and settings show that HaT outperforms state-of-the-art baselines by up to 16.5\% accuracy and saves up to 39\% communication traffic.
\end{abstract}

\section{Introduction}

Deployment of customized learning models presents a critical dilemma. While building specialized models for each user, device, or environment (\textit{domain}) on the edge can yield fine-grained performance and preserve privacy \cite{luspp, kong2023convrelu++}, producing these per-domain models is expensive. Each customized model demands substantial labeling and training efforts, yet in practice, many domains have only scarce labeled data due to prohibitive labeling costs \cite{pmlr-v235-gao24j, dai2024contrastsense}. Moreover, modern learning systems typically serve many such domains (potentially in the hundreds or thousands), in which cases the labeling and retraining overhead becomes unsustainable. 

While one might consider transferring existing models to new domains \cite{pmlr-v235-phan24a, pmlr-v235-lu24i}, significant challenges arise from data and device heterogeneity. Models trained on a particular data distribution may fail to generalize to another, suffering from performance drops or inapplicability to unseen categories. Additionally, resource constraints of different devices, such as memory or computational power, further complicate direct model reuse \cite{li2024flexnn}. Consequently, a more systematic strategy is required to expand existing pool of models to accommodate new target domains without incurring massive cost. In this work, we refer to this process as \textit{learning system expansion}. 

\begin{figure}
    \centering
    \includegraphics[width=1.0\linewidth]{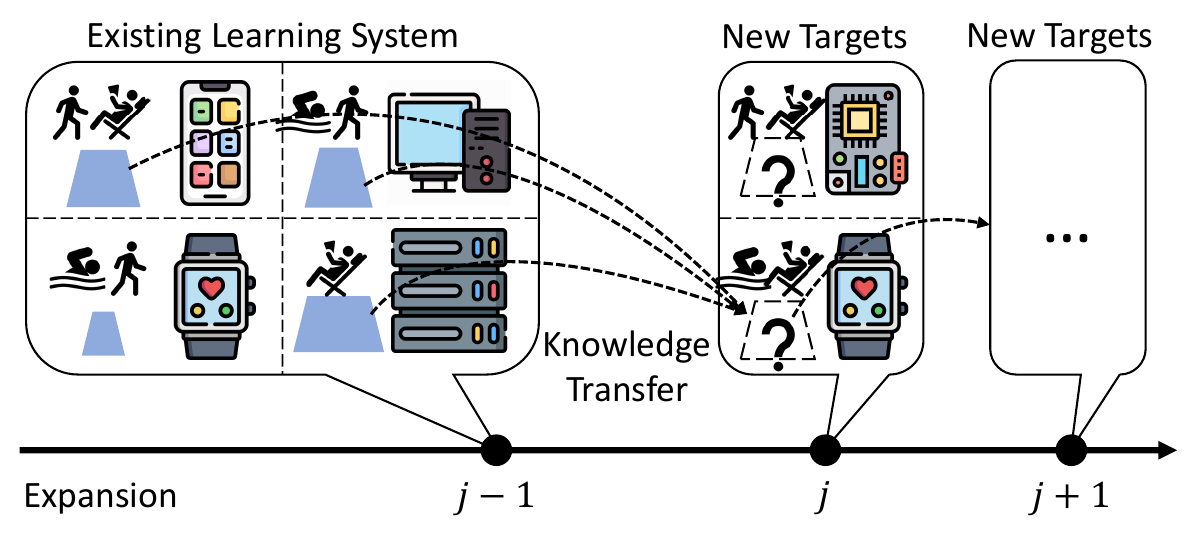}
    \caption{Expanding learning systems is challenging due to label scarcity and large heterogeneity.}
    \label{fig: scenario}
\end{figure}

Figure \ref{fig: scenario} illustrates a realistic scenario for the \textit{learning systems expansion}. Existing domains, such as different users, devices, datasets, or organizations, maintain heterogeneous models for processing local data, referred to as the \textit{source domains}. The new targets, i.e., \textit{target domains}, have limited labeled data but many unlabeled data due to the large labeling overhead. The data within these domains are non-independent and identically distributed (non-IID) with potential shifts in label space. We pose a critical question: \textit{How can we effectively and efficiently expand learning systems?}

\begin{figure*}[!tb]
        \centering

        \subfigure[Potential of knowledge transfer.]{
            \centering
            \includegraphics[width=0.29\linewidth]{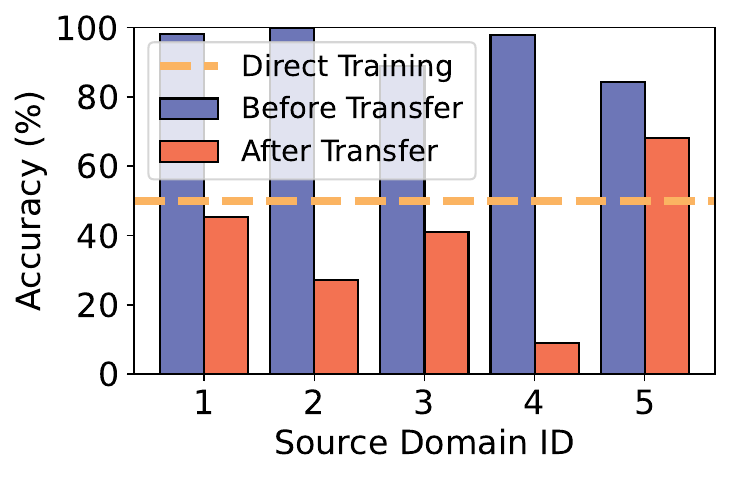}
            \label{fig: data heterogeneity}
            }
        \hspace{0mm}
        \subfigure[Limitations of existing methods.]{
            \centering
            \includegraphics[width=0.31\linewidth]{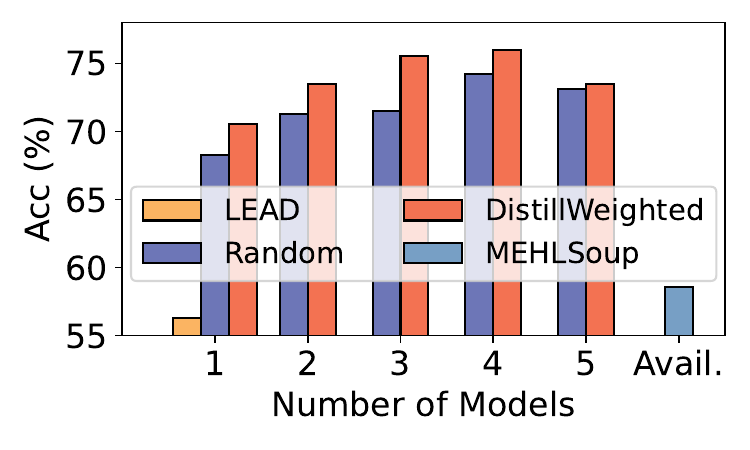}
            \label{fig: num of teachers}
            }
        \hspace{0mm}
        \subfigure[Knowledge conflicts]{
            \vspace{0mm}
            \includegraphics[width=0.28\linewidth]{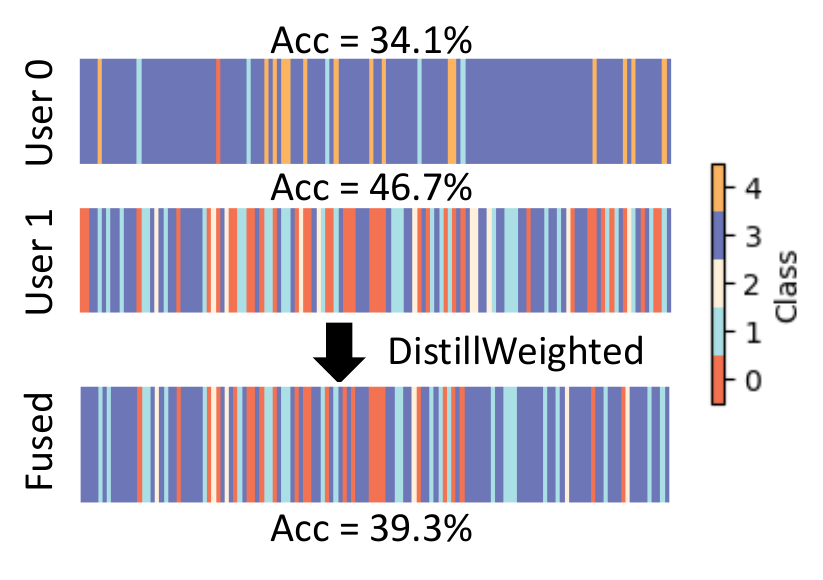}
            \label{fig: conflict}
        }
        
    \caption{Motivation study. (a) When only 10\% data is labeled, direct training is less effective compared with transferring high-quality source models. (b) While leveraging more source models may lead to better performance, existing methods cannot fully exploit the knowledge from source domain models. (c) Knowledge conflicts between source models are hard to be resolved by existing knowledge fusion methods. One bar refers to the prediction result on one sample.} 
\end{figure*}

Existing approaches struggle to handle this question. To handle data heterogeneity, domain adaptation is widely studied to enhance model robustness by aligning feature distributions across domains \cite{wilson2021calda, he2023domain, Qu2024LEAD}. Nevertheless, those works overlook device heterogeneity, making it difficult to adopt existing models across diverse hardware environments. Knowledge distillation addresses device-side constraints by transferring knowledge from teacher models to student models \cite{hinton2015distilling, gou2021knowledge, borup2023distilling, pmlr-v235-peng24a}. Yet existing methods assume that the teacher models are fully reliable and do not account for the impact of data heterogeneity. The non-IID data across domains makes source models less accurate on target domains. Therefore, there is a significant gap in addressing the expansion problem.


To address this gap, we conduct a motivation study and observe that, while each source model is trained on potentially non-IID data, it can still retain valuable “knowledge”—as illustrated by model 5 in Figure \ref{fig: data heterogeneity}—such as the ability to interpret data characteristics and distinguish between classes. Moreover, as shown in Figure \ref{fig: num of teachers}, combining multiple source models can yield more comprehensive knowledge that compensates the weaknesses of individual models. Based on those observations, our key idea is to exploit the high-quality knowledge from multiple source models to train customized target models. 

However, not all knowledge is beneficial for target domains. As shown in Figure \ref{fig: num of teachers}, blindly combining all available models may result in suboptimal performance. Yet identifying high-quality source models is challenging because performance in source domains does not necessarily indicate effectiveness in the target domain, and label scarcity limits direct evaluation. Furthermore, knowledge from different sources may conflict, as illustrated in Figure \ref{fig: conflict}. Resolving these conflicts requires prioritizing specific models, yet the reliability of each model's knowledge can vary dynamically across samples and domains. Additionally, selecting high-quality source models to train target models can be expensive, making optimization essential to reduce overhead. Overall, selectively aggregating and injecting multi-source knowledge into target models while maintaining reasonable overhead is crucial for learning system expansion.

To address these challenges, a practical \textbf{H}eterogeneity-\textbf{a}ware Knowledge \textbf{T}ransfer framework (HaT) is proposed: 1) To identify high-quality source models, HaT first utilizes simple statistical features to filter out low-quality source domains. The remaining models from are then further selected based on their performance on the target domain data. 2) To address knowledge conflicts, an attention-based mixer is trained to assign sample-wise weights to each model's predictions by measuring the representations similarity between models. A knowledge dictionary is constructed to selectively store the fused predictions. This stored knowledge is later injected into the target model, with the transfer speed dynamically adjusted based on the quality of the knowledge. 3) To minimize system overhead, the model selection process is encapsulated in a communication protocol that only transmits models with high potential for further evaluation in the target domains. A low-cost joint training scheme is implemented to simultaneously update the mixer, the target model, and the partially frozen source models.

Extensive experiments have been conducted across various modalities and tasks, including Human Activity Recognition, Gesture Recognition, and Image Classification. The results show that HaT achieves up to 16.5\% higher accuracy, which also reduces communication traffic by up to 39\% and source model execution overhead by 40\%. The key contributions are as follows:

1) We address a practical learning system expansion problem characterized by label scarcity and both data and device heterogeneities.

2) We propose a general system expansion framework, HaT, which efficiently selects, fuses, and injects multi-source knowledge to deliver high-quality customized models for targets with practical system overhead.

3) We evaluate the framework across various tasks, modalities, and architectures, demonstrating superior performance compared with baselines.


\section{Related Works}  

\textbf{Transfer Learning.} Transfer learning explores methods to apply source models for new targets, addressing data or task heterogeneities \cite{pan2009survey, tan2018survey}. In particular, domain adaptation has been extensively studied to align feature distributions between domains \cite{zhu2020deep, wilson2021calda, he2023domain, Qu2024LEAD}. However, these approaches typically require access to both source and target domain data, which may not be feasible. In contrast, test-time adaptation techniques adapt models using only test data, enabling continual learning \cite{gong2024sotta, karmanov2024efficient}. Additionally, multi-source transfer learning methods aim to select source models with better generalizability \cite{tong2021mathematical, agostinelli2022transferability}. Despite these advancements, most approaches fail to address device heterogeneity, which might hinder the direct application of source models to target domains.

\textbf{Knowledge Distillation.}  
In knowledge distillation, an efficient student model is trained using the knowledge from one or more teacher models, such as their predicted pseudo labels or intermediate features \cite{hinton2015distilling, liu2020adaptive, knowledgetransfer2024, pmlr-v235-peng24a}. Specifically, multi-teacher distillation approaches \cite{borup2023distilling, liu2020adaptive, zhang2022confidence} aggregate the knowledge of multiple teachers by assigning weights, aiming to provide the student model with more accurate and comprehensive knowledge. Most knowledge distillation studies focus on a closed-set problem, where high-quality teacher models are predetermined and readily available \cite{hinton2015distilling, liu2020adaptive, zhang2022confidence, borup2023distilling}. However, in the learning system expansion problem, the knowledge from source domain models may not directly transfer to the target domain due to data heterogeneity, leading to suboptimal performance.

\textbf{Model Customization.} Model customization has been extensively studied to meet specific computational and performance requirements \cite{wen2023adaptivenet, li2024flexnn}. Some works explore pre-deployment or post-deployment model generation techniques \cite{cai2020once, wen2023adaptivenet} to search optimal architecture in terms of latency and accuracy. In contrast, HaT emphasizes the knowledge transfer process from the selected source models to any target models that satisfy the customized needs of target domains.

\section{Heterogeneous Learning Systems Expansion}
\subsection{Problem Formulation}
\label{sec: problem formulation}
We define the \textbf{Multi-Round System Expansion (MRSE)} problem to address the ever-growing nature of learning systems. At round \( j \), there are \( N^{S}(j) \) existing source domains, denoted as \( \mathcal{D}^S(j) = \{\mathcal{D}^S_i(j)\}_{i=1}^{N^S(j)} \), and \( N^T(j) \) target domains, \( \mathcal{D}^T(j) = \{\mathcal{D}^T_i(j)\}_{i=1}^{N^T(j)} \). Each target domain in \( \mathcal{D}^T(j) \) requires high-quality, customized models to meet its unique requirements. Once the models for target domains \( \mathcal{D}^T(j) \) are curated, these domains become source domains in the subsequent round: $\mathcal{D}^S(j+1) = \mathcal{D}^S(j) \cup \mathcal{D}^T(j)$. The knowledge in \( \mathcal{D}^S(j+1) \) is then leveraged to curate models for target domains in \( \mathcal{D}^T(j+1) \). The primary objectives are: 1) maximize performance of the curated models on target domainsp; 2) minimize curation overhead, which includes communication and computation costs.

To better understand the process, the MRSE problem can be decomposed into individual \textbf{One-Time System Expansion (OTSE)} problems. For a specific target domain \( \mathcal{D}^T_i(j) = \{X^T_i, Y^T_i, \zeta^T_i\} \), the goal is to curate a model based on the knowledge from source domains \( \mathcal{D}^S(j) = \{X^S_i, Y^S_i, NN^S_i, \zeta^S_i\} \). However, source and target domains exhibit distributional differences between \( X^S_i \) and \( X^T_i \), which hinder the direct applicability of the source model \( NN^S_i = \{f^S_i, g^S_i\} \). The $f^S_i$ and $g^S_i$ are the encoder and the classifier.  The label sets of source domains $Y^S_i$ and target domain $Y^T_i$ may not fully overlap, introducing additional complexity during expansion. Moreover, target domains impose constraints \( \zeta^T_i \), including memory usage and inference speed requirements, which must be considered during model curation. Besides, due to the high cost of labeling, only few (\( \gamma\% \)) data in \( \mathcal{D}^T_i(j) \) is labeled, which is a general assumption to handle potential label space difference.

\subsection{Technical Challenges}
To expand learning systems to new target domains, three key challenges must be addressed:

First, it is difficult to identify high-quality source models within the systems. Estimating performance degradation caused by the data discrepancies based on discrepancy measurements is less practical if requiring access to both source and target domain data \cite{chang2020systematic, Qu2024LEAD}. Additionally, evaluating models solely with few labeled data in the target domains may not accurately reflect the model’s in-the-wild performance, as those data can hardly capture the diversity of the target data.

Second, transferring knowledge from multiple source models to customized target models is challenging due to the knowledge conflicts. Source models make conflicting predictions for the same input, which leads to contradictory update directions and less effective model training. Resolving knowledge conflicts is difficult because the quality and relevance of the knowledge from each model can vary dynamically across different samples. This context-dependent variability makes it challenging to determine which model's knowledge should be prioritized.

Third, minimizing system overhead is crucial during expansion. Evaluating source models on the target domains would incur significant communication and execution costs, given the large number of domains in the systems. However, if all models are not tested in the target domains, high-quality models might be discarded, leading to suboptimal performance. Furthermore, leveraging the knowledge from multiple models to train the target model can also introduce large training overhead, which needs to be minimized.

\section{Design of HaT}

\subsection{Framework Overview}  \label{sec: design}

\begin{figure}[!t]
    \centering
    \includegraphics[width=1.0\linewidth]{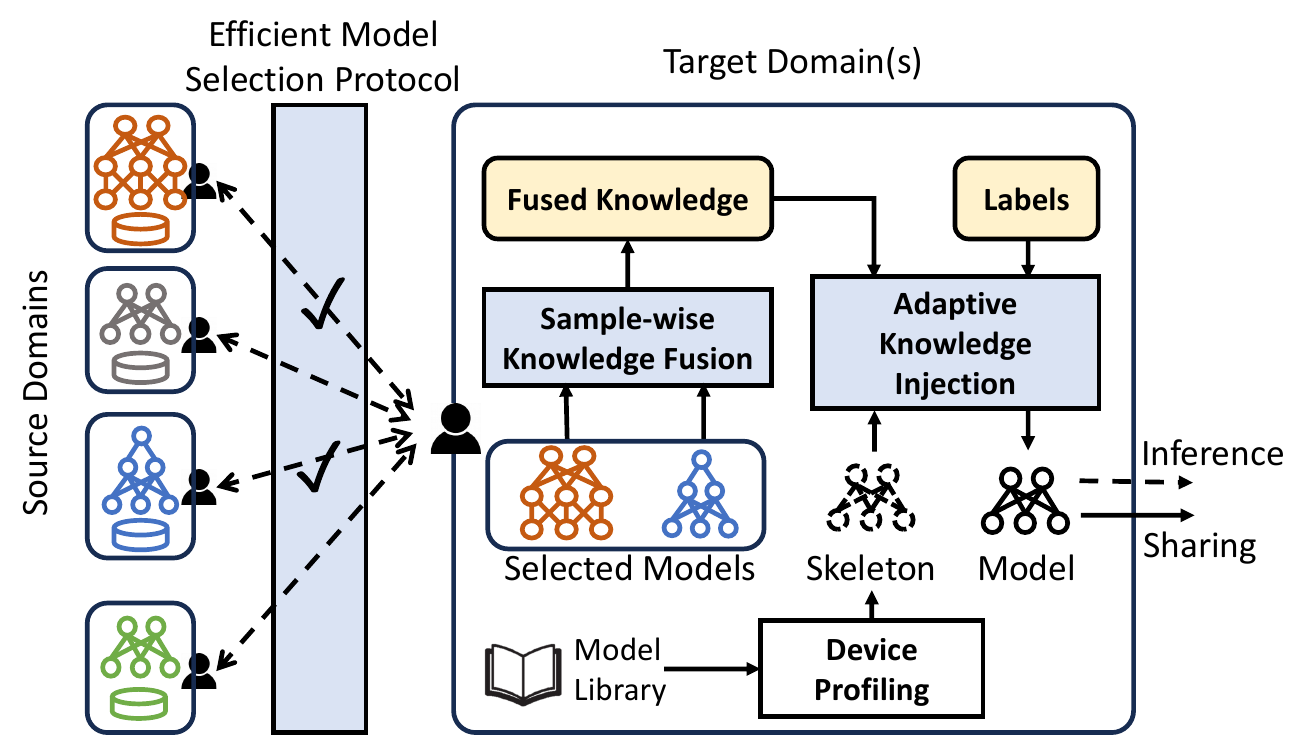}
    \caption{Framework overview of HaT.}
    \label{fig: framework}
\end{figure}

HaT is designed to address the three aforementioned challenges, the process of which is presented in Figure \ref{fig: framework}. HaT first invokes the Efficient Model Selection Protocol to select high-quality source models at a low cost. Later, the Sample-wise Knowledge Fusion is performed to aggregate the conflicting knowledge. Subsequently, the target model is trained with the Adaptive Knowledge Injection based on a low-cost training scheme. We further introduce the details of providing a customized model for one target $\mathcal{D}_t$ (instead of $\mathcal{D}_t^T$ for clarity), the processes of which are scalable and repeatable for any number of targets.
    
\subsection{Efficient Model Selection Protocol}
    The Efficient Model Selection Protocol aims to select high-quality source domain models at a low communication and model inference cost with two stages of selection:
    
\textbf{Feature-based Coarse Selection.}
    Lightweight features $l=\{l(n)\}$, such as mean value, standard deviation, skewness, and kurtosis, are extracted from both the source and target domain data. These features provide a coarse-grained description of domain characteristics without relying on any learning models. The target domain transmits its extracted features to the source domains, where each source domain $i$ computes a feature similarity score: $s_i =\sum_{n=1}^N sim(l^S_i(n), l_t(n))/N,$
    where $N$ is the total number of features. The similarity function $sim(\cdot)$ is set to cosine similarity. The target domain then selects the top $\eta\%$ of source domains with the highest similarity scores. Only the selected domains that receive a model inquiry from the target domain transmit their models back. 
    

\textbf{Centroids-Accuracy Joint Selection.}
    After the models are received in the target domain, the accuracy $Acc_i$ of the $i$-th source model is computed on the target domain labeled data, which, however, can be unreliable due to label scarcity. To address this, we further use class centroids, which represent the class-wise feature distributions captured by the models. These centroids are computed by averaging the encoder outputs for samples within each class:
    \begin{equation}
    \begin{split}
        c_{t,m} = \frac{1}{K}\sum_{k=1}^K{h_i(k)}, k \in \{k|max(g^S_i(h_i(k))) = m \\
        \text{and } rank(e(k))<\omega 
        \}  
    \end{split}
    \end{equation}
    where $h_i(k) = f^S_i(x(k))$ represents the features extracted by the encoder $f^S_i$ of the $i$-th domain from the $k$-th data sample. The classifier output, $g^S_i(h_i(k))$, is processed with $max(\cdot)$ to obtain a pseudo label, which may be inaccurate due to data heterogeneity. To enhance the quality of the centroids, entropy, a measure of the prediction confidence, is used to filter out features with low certainty. The entropy $e(k)$ of the logits output $g^S_i(h_i(k))$ is calculated, and the $\omega=75\%$ of features with the lowest entropy (i.e., the highest confidence) are selected for centroid extraction.

    The centroids of $\mathcal{D}_t$ and $\mathcal{D}^S_i$ are compared to estimate the data distribution similarity $s^{fine}(t, i)$: $s^{fine}(t, i) =\sum_{m=1}^M \text{sim}(c_{t,m}, c^S_{i,m})/M,$ where $M$ is the number of overlapping classes between $\mathcal{D}_t$ and $\mathcal{D}^S_i$. A higher value of $s^{fine}(t, i)$ indicates a greater similarity in data distribution, suggesting that the source model is likely to experience a smaller performance drop when applied to the target data. The centroid similarity score $s^{fine}(t, i)$ is then multiplied by the accuracy $Acc_i$ to rank the source domains and determine the final set of $N_p$ selected models. 


    This approach reduces communication traffic and inference costs, as only a subset of models is transmitted and tested in $\mathcal{D}_t$, given that features are much smaller in size compared to models. Compared to methods that require access to each sample in $\mathcal{D}_t$ and $\mathcal{D}^S_i$ \cite{he2023domain, du2024domain}, HaT enhances privacy by exchanging only high-level features and models between domains. Furthermore, privacy can be further strengthened by selectively sharing non-sensitive features while excluding sensitive content \cite{Qu2024LEAD}.
    
\subsection{Sample-wise Knowledge Fusion}
To resolve knowledge conflicts, an attention-based mixer is trained to assign sample-wise weights. Simultaneously, a cost-effective adaptation is applied to the source models to enhance their knowledge quality.

\textbf{Attention-based Mixer.} 
The attention-based mixer aggregates conflicting predictions from the selected models by leveraging the sample-wise feature adjacency between the source and target models. The feature $h_t(k)$ extracted by the target encoder is projected through a linear layer $L^{\text{query}}$ to obtain the query vector $q(k)$. Similarly, the features $h_i(k)$, extracted by the selected source models, are projected through the respective linear layers $L_i^{\text{key}}$ to obtain the key vectors $key_i(k)$:
\begin{align}
    q(k) &= L^{\text{query}}(h_t(k)), \\
    key_i(k) &= L_i^{\text{key}}(h_i(k)).
\end{align}
Different from the classic attention mechanism \cite{vaswani2017attention}, which uses a single linear layer to compute the keys, the mixer utilizes multiple linear layers $L_i^{\text{key}}$ to accommodate the heterogeneous source model architectures. Since models from different source domains extract features of varying dimensions, the input size of each $L_i^{\text{key}}$ must be tailored accordingly. The output size of $L_i^{\text{key}}$ is standardized to a common dimension for subsequent computations. The similarities between the query and the keys are then calculated and normalized using SoftMax to obtain the attention score $w_i(k)$ for the $i$-th model on the data sample $x(k)$:
\begin{equation}
    w_i(k) = \text{SoftMax}(q(k) \cdot key_i(k)), \quad i = 1, 2, \ldots, N_p
\end{equation}
The attention score $w_i(k)$ measures the feature similarity between the target and the selected models, which is used to aggregate the predictions from the selected classifiers $g_i(k)$:
\begin{equation} \label{equ: fused prediction}
    p^{\text{mix}}(k) = \sum_i w_i(k) \cdot Map[g^S_i(h_i(k))],
\end{equation}
where $Map[\cdot]$ projects the label spaces of different domains to a unified label space that includes all categories. If the features extracted by the target model and the $i$-th model are highly similar, a higher weight $w_i(k)$ is assigned to the prediction of the $i$-th model. This is because the $i$-th classifier is likely to be more accurate on data from a distribution that closely resembles the data it is trained with. However, training the attention-based mixer is challenging, as it depends on input from the target model, which itself requires training. To address this, a low-cost joint training scheme is proposed to train the mixer and the target model simultaneously, which is detailed in Section \ref{sec: knowledge inject}.

\begin{figure}
    \centering
    \includegraphics[width=0.85\linewidth]{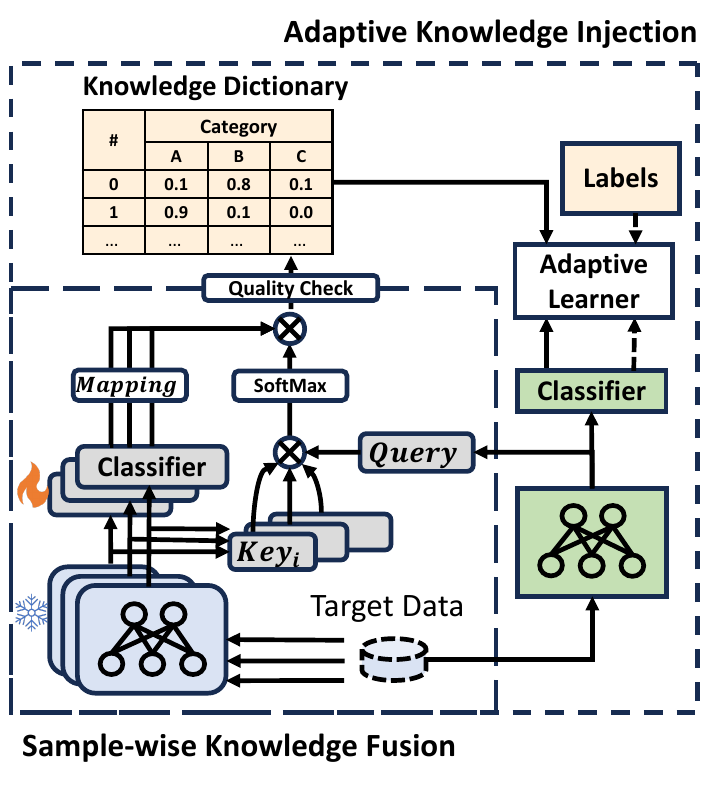}
    \caption{Details of the Sample-wise Knowledge Fusion and the Adaptive Knowledge Injection process.}
    \label{fig: knowledge distillation}
\end{figure}

\textbf{Cost-effective Adaptation}
To further improve the accuracy of the fused predictions, the classifiers are trained jointly with the mixer, while their encoders remain frozen, since adapting all $N_p$ selected source models would be computationally expensive. This approach reduces the computational burden, as classifiers are typically lightweight \cite{he2016deep}. Early-stage experiments indicate that adapting the classifiers alone is sufficient to provide high-quality predictions for the attention-based mixer. Additionally, freezing the encoders accelerates the knowledge aggregation process. By precomputing features for all target domain data using the frozen encoders and storing them in memory, they are ready to be fetched when the mixer requires the prediction results for a sample $x(k)$. This process eliminates the need to repeatedly execute the forward pass of the selected encoders, thereby reducing the overall computation time.

\subsection{Adaptive Knowledge Injection} 
\label{sec: knowledge inject} 

Considering the constraints $\zeta_t$ on memory and the inference speed of the target, HaT selects a suitable model skeleton from a model library through device profiling. The model library includes open-source models from the Internet, the architectures from $\mathcal{D}^S$, or models contributed by the target. Then, the fused knowledge is distilled into the target model:
\begin{equation} \label{equ: adaptive loss}
    L_{\text{ada}} = L^{\text{label}} + \alpha \, L^{\text{distill}}(g_t \cdot f_t(x(k)), KD(k)),
\end{equation}
where $L^{\text{label}}$ represents the cross-entropy loss on the labeled data, and $L^{\text{distill}}$ denotes the distillation loss based on the pseudo labels. A knowledge dictionary $KD$ and an adaptive learner are further designed to enhance the training efficacy.

\textbf{Knowledge Dictionary.}
Directly learning from the fused prediction $p^{\text{mix}}$ may hinder the convergence of the target model. This is because the aggregation results from the mixer change dynamically during its training (e.g., the pseudo-label of a sample may shift from class A to C), leading to conflicting gradient update directions. Additionally, the quality of the fused predictions may fluctuate from epoch to epoch. The knowledge dictionary $KD$ is introduced to provide more stable learning objectives. After each model update, the fused predictions from the attention-based mixer are stored in the $KD$ only if the accuracy of the mixer improves. These predictions are stored in a soft-label format rather than as one-hot vectors, allowing the target model to capture the confidence levels of the mixer in its fused predictions. In subsequent epochs, the target model learns from the soft pseudo labels in the $KD$.

\textbf{Adaptive Learner.}
Given the varying quality of the fused predictions in the knowledge dictionary, an adaptive learner is employed to adjust the weight $\alpha$ of the distillation loss: $\alpha = m(Acc_{\text{train}} - b)$, where $Acc_{\text{train}}$ represents the accuracy of the attention-based mixer on the training data. The $m$ and $b$ are predetermined hyperparameters. The $m$ controls the scaling factor of the weight $\alpha$, while $b$ serves as a threshold to prevent the target model from learning from fused predictions of low quality. The weight $\alpha$ increases when the fused prediction accuracy is high, allowing the model to learn more effectively from reliable predictions. 

\textbf{Low-cost Joint Training} \label{sec: joint training} 
The training of the attention-based mixer, the unfrozen source classifiers, and the target domain model is complex, as they depend on inputs from each other. To enable a cost-effective training process, a joint training scheme is developed as shown in Algorithm \ref{algo: joint training}. 

\begin{algorithm}[!t]
\caption{Low-cost Joint Training of HaT}\label{algo: joint training}
\begin{algorithmic}[1]
    
    \STATE \textbf{Input:} Target labeled and unlabeled data $X^{\{l, u\}}$, labels $Y^{l}$, selected models $\{f^S_i\}_{i=1}^{N_p}, \{g^S_i\}_{i=1}^{N_p},$
    \STATE \textbf{Output:} 
    Target encoder and classifier $f_t, g_t$
    \STATE Initialize $g_t, f_t, Mixer, KD$
    \STATE $\{h_i^{\{l, u\}}\}$ = Encoding($X^{\{l, u\}}$, $\{f^S_i\}_{i=1}^{N_p}$)
    \FOR{epoch in Epochs}
        \STATE $h_t^{l}, h_t^{u}=f_t(X^{l}), f_t(Sampling(X^{u}))$
        \STATE 
        $Mixer$ = Mixer\_Update($\{h_i^{l}\}, h_t^{l}, Mixer, Y^{l}, \{g^S_i\}$)
        \STATE $f_t, g_t$ = Model\_Update($f_t, g_t, h_t^{\{l, u\}}, Y^{l}, KD$) \\
        \IF{Quality\_Improve($Mixer$)}
            \STATE $KD$=Dict\_Update($KD, Mixer, \{h_i^{u}\}, h_t^{u}$)
        \ENDIF
    \ENDFOR
\end{algorithmic}
\end{algorithm}
The labeled and unlabeled data, $X^{\{l,u\}}$, are encoded by the frozen source encoders $\{f^S_i\}_{i=1}^{N_p}$ to high-level features, which are kept for later training. In each epoch, the $f_t$ encodes $X^{\{l,u\}}$, generating representations $h_t^{l}$ and $h_t^{u}$. To manage the computational overhead of processing a large volume of unlabeled data $X^u$, only a subset of $X^u$ is randomly sampled in each epoch, with the sample size kept proportional to the size of the labeled data. This strategy ensures that the entire set of unlabeled data is progressively utilized over multiple iterations, thereby reducing training time and memory usage of each epoch without compromising model performance. The $KD$ is updated only when the mixer's quality improves, minimizing the cost of generating pseudo-labels for all unlabeled data. 

The cross-entropy loss is computed using $p^{\text{mix}}$ and labels $Y^{l}$ and minimized by one optimizer to train the mixer and the unfrozen classifiers (illustrated in gray in Figure \ref{fig: knowledge distillation}). Equation (\ref{equ: adaptive loss}) is minimized by a separate optimizer to train the target model (illustrated in green in Figure \ref{fig: knowledge distillation}). After the model training, only $g_t$ and $f_t$ are stored for the inference.


\section{Evaluations} \label{sec: exp public}
\subsection{Experiment Setting} \label{sec: experiment setting}
\textbf{Datasets} HaT is evaluated on four datasets, HARBox\cite{ouyang2021clusterfl}, ImageNet-R\cite{hendrycks2021many}, NinaPro\cite{pizzolato2017comparison}, and Alzheimer's Disease (AD)\cite{ouyang2023harmony} that span six modalities, three tasks, and different scales. For each dataset, six different model architectures are included as model libraries, with further details introduced in Appendix \ref{appendix: datasets} and \ref{app: model library}.

\textbf{Baselines} The five most relevant baselines from knowledge distillation, model aggregation, and domain adaptation are implemented and slightly adapted for comparison, including DistillWeighted \cite{borup2023distilling}, DistillNearest \cite{borup2023distilling}, LEAD \cite{Qu2024LEAD}, and MEHLSoup \cite{li2024leaerning}. We also propose a baseline called AccDistill, which distills the knowledge from ensemble models based on selected models from the source domains. The details of baselines are further introduced in Appendix \ref{appendix: baselines}.


\textbf{Real-world Testbed.} The system is deployed on a server and an edge device, the Nvidia Jetson Xavier. To simulate different source domains, the data and models are stored in separate folders on the server due to the limited number of available devices. The target domain data is deployed on the edge device. The model training overhead, including time and memory usage, is measured on the edge devices, which are closely correlated with energy consumption. The communication overhead is monitored by tracking the network traffic between the server and the edge device.

\begin{table*}[!htbp]
\caption{
Performance comparison in the MRSE setting. Accuracy refers to the average accuracy of four rounds of expansion. Traffic is the total communication cost during expansion.}
\label{tab: system expansion}

\centering
\begin{tabular}{c|cc|cc|cc|cc}
\toprule
\multirow{2}{*}{Methods} & \multicolumn{2}{c|}{HARBox} & \multicolumn{2}{c|}{ImageNet-R} & \multicolumn{2}{c|}{NinaPro} & \multicolumn{2}{c}{AD} \\ 
& Accuracy & Traffic & Accuracy & Traffic & Accuracy & Traffic & Accuracy & Traffic \\ 
\midrule
LEAD & 51.46 & 1483 & 48.17 & 508 & 44.94 & 2.17 & 36.46 & 202 \\ 
MEHLSoup & 62.98 & 1483 & 47.57 & 508 & 43.32 & 2.17 & 31.04 & 202 \\ \midrule
AccDistill & 73.40 & 4034 & 57.56 & 1786 & 35.65 & 11.8 & 52.71 & 768 \\
DistillNearest & 74.95 & 4034 & 57.64 & 1786 & 40.75 & 11.8 & 58.12 & 768 \\ 
DistillWeighted & 75.42 & 4034 & 57.66 & 1786 & 41.02 & 11.8 & 56.46 & 768 \\ \midrule
HaT & \textbf{79.27} & \textbf{2517} & \textbf{59.30} & \textbf{1279} & \textbf{45.12} & \textbf{6.95} & \textbf{63.96} & \textbf{525} \\ 
\bottomrule
\end{tabular}
\end{table*}

\textbf{Implementation Details.} 
The learning rates of target models and mixer are searched among $\{\text{5e-4, 1e-3, 5e-3, 1e-2}\}$ for different datasets. Training epochs of both are 200 and 100. The scaling ratio $m$ and the bias $b$ are determined using a grid search within the ranges $[1.0, 4.0]$ and $[0, 0.5]$, with step sizes of 0.5 and 0.1, respectively. The $N_p$ is set to three. A sensitivity analysis is provided in Appendix \ref{app: Sensitivity Analysis}.

\subsection{Result Comparison in MRSE} \label{sec: expansion}
We present the effectiveness of HaT in the MRSE setting. The domains in the datasets are split into four groups to simulate the learning systems expansion process in section \ref{sec: problem formulation}. The details of the data split are in Appendix \ref{app: data splits MRSE}.

\begin{figure}[!t]
    \centering
    \includegraphics[width=0.95\linewidth]{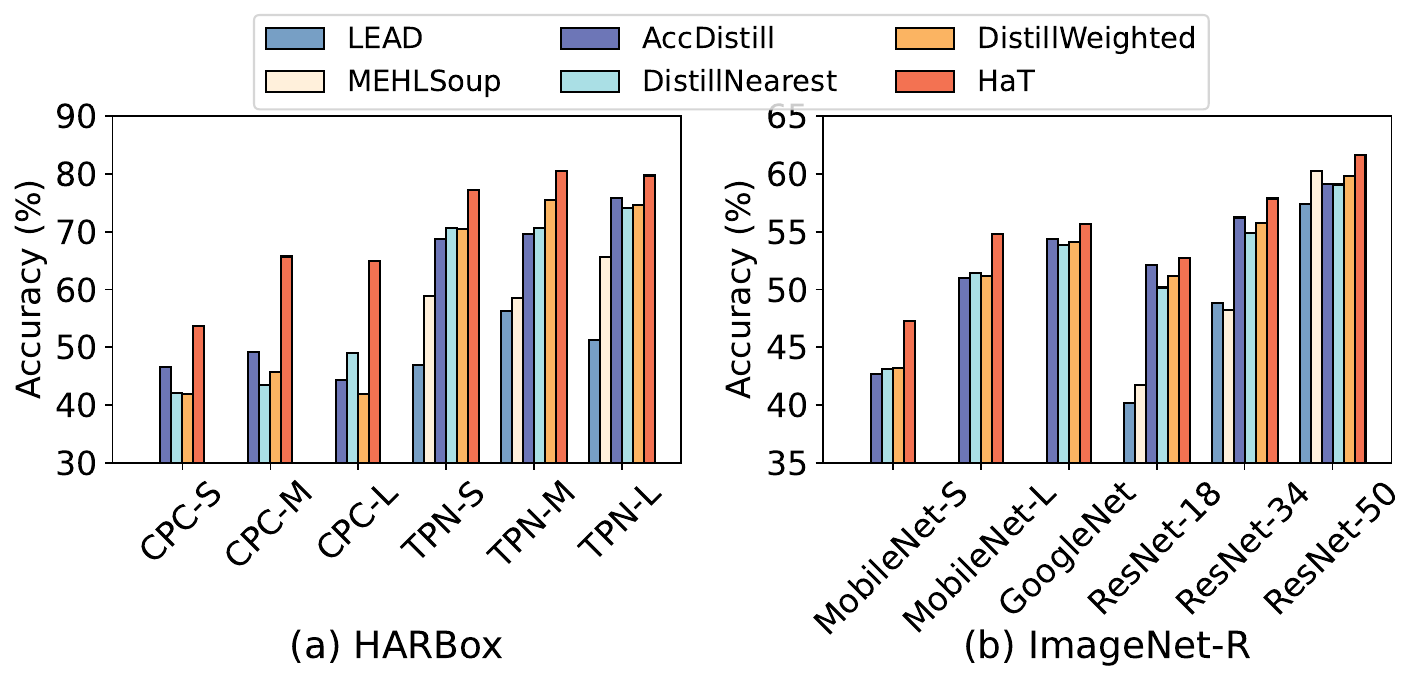}
    \caption{Performance comparison with different target models.}
    \label{fig: varied archi}
\end{figure}


Table \ref{tab: system expansion} presents the performance across different rounds of expansion, with more detailed results provided in Appendix \ref{app: multi-round results MRSE}. Compared to baselines that either leverage limited knowledge from source domains (LEAD and MEHLSoup) or transfer knowledge in a static manner (AccDistill, DistillNearest, and DistillWeighted), HaT delivers more effective customized models for target domains. This improvement is achieved by fusing knowledge from multiple high-quality models at a finer granularity and dynamically injecting knowledge during training. Furthermore, since models trained in earlier rounds are incorporated into subsequent rounds, HaT enables a progressive transfer of higher-quality knowledge, resulting in consistently improved performance for new target domains.

In Table \ref{tab: system expansion}, as the system scales, the increase in the number of source domains results in larger traffic. HaT effectively reduces this communication burden across all datasets. Note that the communication traffic of LEAP and MEHLSoup is not directly comparable to HaT, as these methods are constrained to using a limited subset of source domains with architectures that align with the target domains. This limitation stems from their inability to handle model heterogeneity, which also contributes to their inferior performance. Besides, HaT reduces the model inference cost during selection by 40\%, since only a portion of source models are transmitted and executed on the target domains. Overall, the results demonstrate HaT's effectiveness and efficiency in expanding learning systems.
\begin{figure}[!t]
    \centering
    \includegraphics[width=0.95\linewidth]{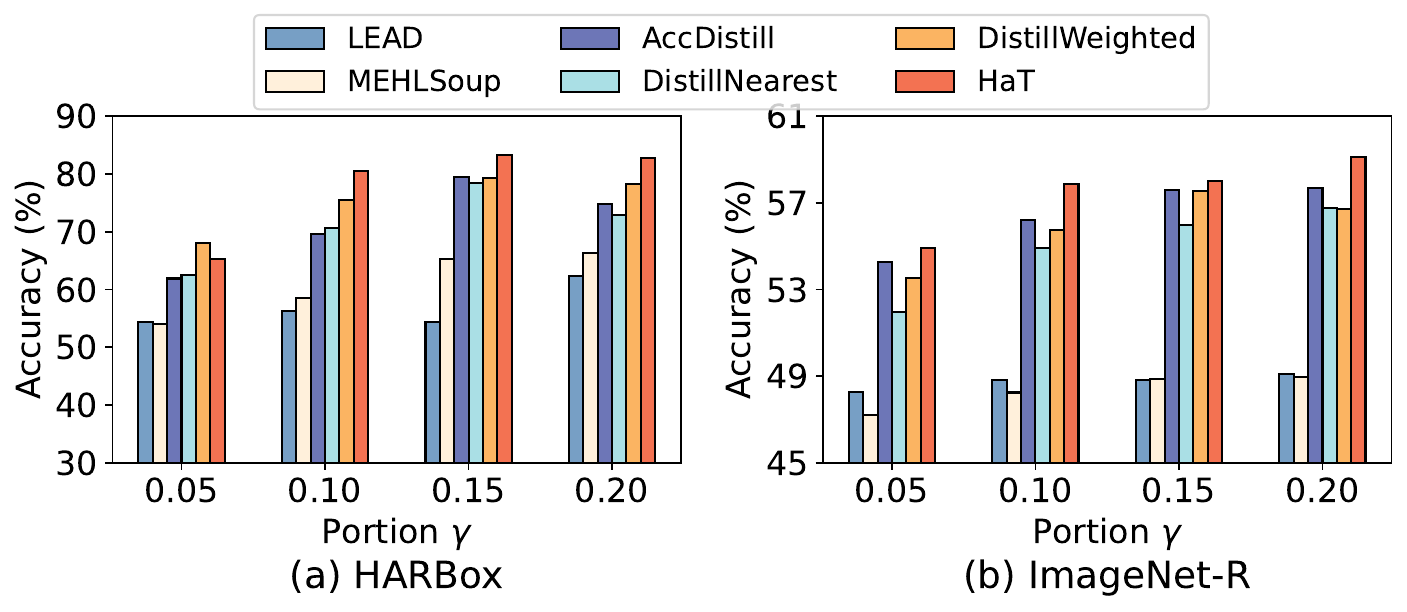}
    \caption{Performance comparison with varied $\gamma$.}
    \label{fig: varied portion}
\end{figure}
\subsection{Result Comparisons in OTSE} \label{sec: lodo}
We further present the performance of HaT on HARBox and ImageNet-R in the OTSE setting. The preprocessing details are in Appendix \ref{app: data splits OTSE}. More results on the other two datasets can be found in Appendix \ref{app: additional results OTSE}

\textbf{Vaired Target Architectures.}
Figure \ref{fig: varied archi} presents the results when different architectures are selected as the target models. Unlike domain adaptation and model merging methods that are limited to source models matching the target architecture (this makes them inapplicable when the required architecture, such as CPC-S for HARBox, is not available among the source models), HaT effectively utilize knowledge from source models with diverse architectures, thus achieving superior accuracy for most architectures. These results demonstrate HaT’s versatility in delivering high-quality customized models across diverse tasks and modalities.

\textbf{Varied Portion of Labeled Data.}  
As shown in Figure \ref{fig: varied portion}, HaT outperforms or achieves comparable performance to the baselines in most cases when the portion $\gamma$ is varied. This is achieved by adaptively fusing and injecting high-quality knowledge from the selected source models. In few cases, the best baseline slightly outperforms HaT. This may be due to the use of a fixed threshold $b$ in the adaptive learners, which may not yield an optimal weight $\alpha$ across different $\gamma$. We plan to explore dynamic thresholding to further enhance HaT's robustness. Besides, we observe that the performance of the baselines is generally lower than reported in their original papers, which is likely due to the more challenging settings involving labels scarcity and large heterogeneities that HaT addresses.


\textbf{Varied Number of Selected Models.} Figure \ref{fig: selection number}(a) shows that HaT consistently achieve better performance when $N_p$ varies, which highlights the effectiveness of HaT in fusing knowledge from multiple models. As $N_p$ increases, the accuracy first increases, which justifies the usage of multiple source models in HaT. When $N_p$ continues to increase, the accuracy drops. It might be due to the low quality knowledge in the additional models, which shows the importance of source models selection.

\subsection{Ablation Study}

\begin{table}[!tb]
\caption{Ablation study in OTSE. SwKF and AKI refer to Sample-wise Knowledge Fusion and Adaptive Knowledge Injection.}
    \label{tab: ablation}
    \scalebox{0.88}{
    \begin{tabular}{c|cccc}
    \toprule
    Design & HARBox   & ImageNet-R     & NinaPro   & AD     \\ \midrule
    HaT & \textbf{80.57}  & \textbf{57.87} & \textbf{49.38} & \textbf{67.33}\\ \midrule
    w/o FbCS & 73.86  & 56.86 & 47.75 &  63.83  \\
    w/o CAJS & 79.92  & 56.45  & 44.95 & 67.17 \\ \midrule
    w/o SwKF & 69.05 & 56.14 & 41.75 & 60.17  \\
    w/o AKI & 66.73 & 53.29 & 40.66 & 48.50 \\  
    \bottomrule
    \end{tabular}
    }
\end{table}

\textbf{Design Effectiveness.} As shown in Table \ref{tab: ablation}, the Feature-based Coarse Selection and Centroids-Accuracy Joint Selection enhance performance by leveraging the statistical features and high-level representations that accurately reflect the domain similarity and the source models effectiveness. The combination of both selection stages demonstrates stronger generalizability across datasets. The Sample-wise Knowledge Fusion achieves an 11.5\% accuracy improvement on HARBox, which is attributed to the sample-wise weights learned by the attention-based mixer could more effectively combine predictions from the source models. Additionally, the accuracy improvement of Adaptive Knowledge Injection highlights the benefit of selectively storing fused predictions and adjusting the weight of the distillation loss based on their quality.


\begin{figure}[!tb]
    \centering
    \includegraphics[width=1.0\linewidth]{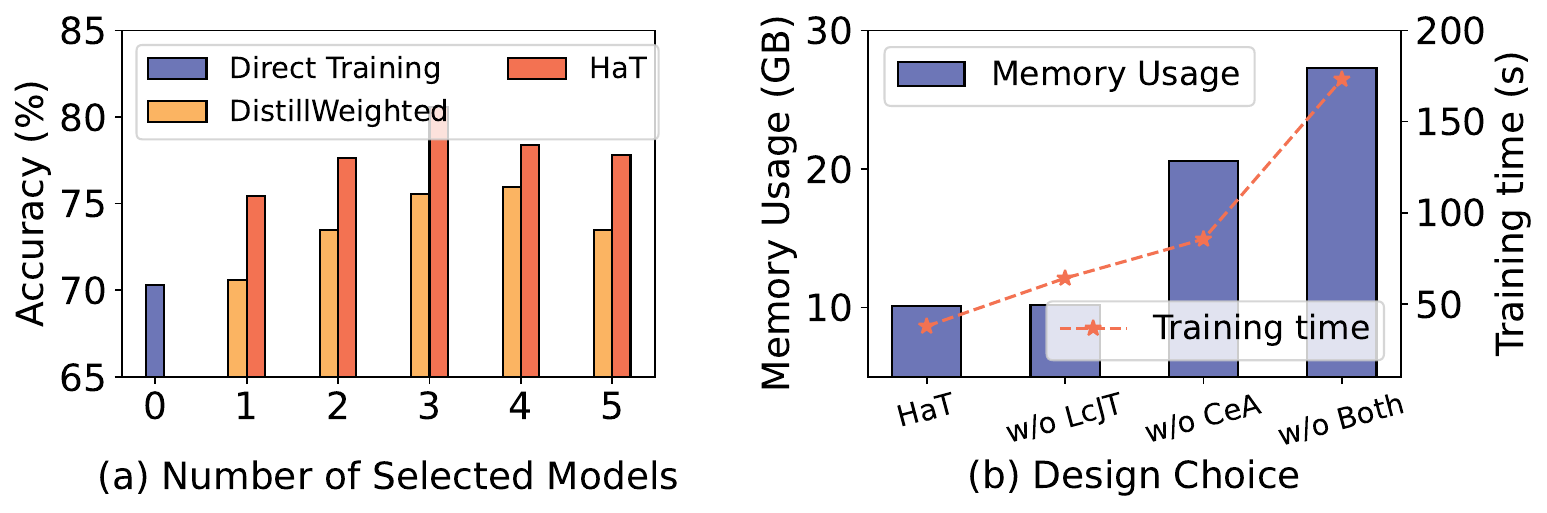}
    \caption{(a) The impact of the number of selected models. (b) The impact of the Cost-effective Adaptation (CeA) and Low-cost Joint Training (LcJT).}
    \label{fig: selection number}
\end{figure}
\begin{table}[!t]
    \centering
    \begin{center}
    
      \caption{Performance comparison of different model selection methods on the HARBox dataset.} \label{tab: selection methods}
      
      \begin{tabular}{c|c|cc|c}
        \toprule
                 & Random & Accuracy & PARC & HaT  \\\midrule
        Acc (\%) & 56.21  & 58.99    & 64.59 & \textbf{65.70} \\
        Traffic (MB) & 13.0  & 858    & 858 & \textbf{526} \\
        \bottomrule
      \end{tabular}
      
    \end{center}
\end{table}

\begin{table}[!t]
\centering
\caption{
Performance comparison of different knowledge fusion methods on the HARBox dataset.} \label{tab: accuracy of fused predictions}
\begin{threeparttable}
    \begin{tabular}{c|c|cc|c}
    \toprule
    Metrics & Nearest  & Equal  & Weighted & HaT  \\\midrule
    $\text{P-Acc}^*$ (\%) & 49.22 & 46.08  & 50.93 & \textbf{75.01} \\
    Acc (\%) & 43.48 & 46.81  & 45.63 & \textbf{65.70} \\
    \bottomrule
    \end{tabular}
\begin{tablenotes}\footnotesize
        \item[*] The accuracy of the pseudo labels.
  \end{tablenotes}
\end{threeparttable}
\end{table}

\textbf{Optimizing the Training Overhead.}  
Figure \ref{fig: selection number}(b) presents the training overhead on the ImageNet-R dataset. The Cost-effective Adaptation, which partially tune the selected models during training, lead to a 2.0$\times$ reduction in memory usage and a 2.3$\times$ reduction in training time due to fewer parameters being optimized. Similarly, incorporating Low-cost Joint Training reduces the per-epoch training time from 64.1s to 37.8s. Overall, HaT achieves significant reductions in both training time (4.6$\times$) and memory usage (2.7$\times$), indicating a more energy-efficient training process.


\textbf{Alternatives in Model Selection.} Several alternative model selection methods are compared with the selection approach of HaT in Table \ref{tab: selection methods}. The knowledge transfer process in HaT is applied to all selection methods. Accuracy and the PARC criteria achieve better performance compared with random selection. However, both methods rely on the labeled data, making them less effective when presented with label scarcity. In contrast, the Efficient Model Selection Protocol in HaT leverages both labeled and unlabeled data for selection and avoids full model transmission, resulting in a 1.1\% accuracy improvement while using only 61.3\% traffic of the communication expense.

\textbf{Alternatives in Knowledge Fusion.} Different knowledge fusion methods are compared in Table \ref{tab: accuracy of fused predictions}. \textit{Nearest} represents only one model is selected and used. \textit{Weighted} indicates the use of the fusion method from DistillWeighted. \textit{Equal} refers to assigning equal weights to all selected models. The accuracy of the pseudo labels generated by HaT is 11.6\% higher than the best alternative method. Consequently, by learning from the higher-quality fused knowledge, the target models achieve a 16.5\% improvement in accuracy.



\bibliography{reference}

\begin{thebibliography}{45}
\providecommand{\natexlab}[1]{#1}
\providecommand{\url}[1]{\texttt{#1}}
\expandafter\ifx\csname urlstyle\endcsname\relax
  \providecommand{\doi}[1]{doi: #1}\else
  \providecommand{\doi}{doi: \begingroup \urlstyle{rm}\Url}\fi

\bibitem[Agostinelli et~al.(2022)Agostinelli, Uijlings, Mensink, and Ferrari]{agostinelli2022transferability}
Agostinelli, A., Uijlings, J., Mensink, T., and Ferrari, V.
\newblock Transferability metrics for selecting source model ensembles.
\newblock In \emph{Proceedings of the IEEE/CVF Conference on Computer Vision and Pattern Recognition}, pp.\  7936--7946, 2022.

\bibitem[Bolya et~al.(2021)Bolya, Mittapalli, and Hoffman]{bolya2021scalable}
Bolya, D., Mittapalli, R., and Hoffman, J.
\newblock Scalable diverse model selection for accessible transfer learning.
\newblock \emph{Advances in Neural Information Processing Systems}, 34:\penalty0 19301--19312, 2021.

\bibitem[Borup et~al.(2023)Borup, Phoo, and Hariharan]{borup2023distilling}
Borup, K., Phoo, C.~P., and Hariharan, B.
\newblock Distilling from similar tasks for transfer learning on a budget.
\newblock In \emph{Proceedings of the IEEE/CVF International Conference on Computer Vision}, pp.\  11431--11441, 2023.

\bibitem[Cai et~al.(2020)Cai, Gan, Wang, Zhang, and Han]{cai2020once}
Cai, H., Gan, C., Wang, T., Zhang, Z., and Han, S.
\newblock Once for all: Train one network and specialize it for efficient deployment.
\newblock In \emph{International Conference on Learning Representations}, 2020.

\bibitem[Chang et~al.(2020)Chang, Mathur, Isopoussu, Song, and Kawsar]{chang2020systematic}
Chang, Y., Mathur, A., Isopoussu, A., Song, J., and Kawsar, F.
\newblock A systematic study of unsupervised domain adaptation for robust human-activity recognition.
\newblock \emph{Proceedings of the ACM on Interactive, Mobile, Wearable and Ubiquitous Technologies}, 4\penalty0 (1):\penalty0 1--30, 2020.

\bibitem[C{\^o}t{\'e}-Allard et~al.(2019)C{\^o}t{\'e}-Allard, Fall, Drouin, Campeau-Lecours, Gosselin, Glette, Laviolette, and Gosselin]{cote2019deep}
C{\^o}t{\'e}-Allard, U., Fall, C.~L., Drouin, A., Campeau-Lecours, A., Gosselin, C., Glette, K., Laviolette, F., and Gosselin, B.
\newblock Deep learning for electromyographic hand gesture signal classification using transfer learning.
\newblock \emph{IEEE transactions on neural systems and rehabilitation engineering}, 27\penalty0 (4):\penalty0 760--771, 2019.

\bibitem[Dai et~al.(2024)Dai, Xu, Yoon, Li, Tan, and Lee]{dai2024contrastsense}
Dai, G., Xu, H., Yoon, H., Li, M., Tan, R., and Lee, S.-J.
\newblock Contrastsense: Domain-invariant contrastive learning for in-the-wild wearable sensing.
\newblock \emph{Proceedings of the ACM on Interactive, Mobile, Wearable and Ubiquitous Technologies}, 8\penalty0 (4):\penalty0 1--32, 2024.

\bibitem[Du et~al.(2024)Du, Li, Li, Lu, Zhu, and Li]{du2024domain}
Du, Z., Li, X., Li, F., Lu, K., Zhu, L., and Li, J.
\newblock Domain-agnostic mutual prompting for unsupervised domain adaptation.
\newblock In \emph{Proceedings of the IEEE/CVF Conference on Computer Vision and Pattern Recognition}, pp.\  23375--23384, 2024.

\bibitem[Gao et~al.(2024)Gao, Qian, Ni, Gan, Hasegawa-Johnson, Chang, and Zhang]{pmlr-v235-gao24j}
Gao, H., Qian, K., Ni, J., Gan, C., Hasegawa-Johnson, M.~A., Chang, S., and Zhang, Y.
\newblock Speech self-supervised learning using diffusion model synthetic data.
\newblock In Salakhutdinov, R., Kolter, Z., Heller, K., Weller, A., Oliver, N., Scarlett, J., and Berkenkamp, F. (eds.), \emph{Proceedings of the 41st International Conference on Machine Learning}, volume 235 of \emph{Proceedings of Machine Learning Research}, pp.\  14790--14810. PMLR, 21--27 Jul 2024.

\bibitem[Gong et~al.(2024)Gong, Kim, Lee, Chottananurak, and Lee]{gong2024sotta}
Gong, T., Kim, Y., Lee, T., Chottananurak, S., and Lee, S.-J.
\newblock Sotta: Robust test-time adaptation on noisy data streams.
\newblock \emph{Advances in Neural Information Processing Systems}, 36, 2024.

\bibitem[Gou et~al.(2021)Gou, Yu, Maybank, and Tao]{gou2021knowledge}
Gou, J., Yu, B., Maybank, S.~J., and Tao, D.
\newblock Knowledge distillation: A survey.
\newblock \emph{International Journal of Computer Vision}, 129\penalty0 (6):\penalty0 1789--1819, 2021.

\bibitem[Haresamudram et~al.(2021)Haresamudram, Essa, and Pl{\"o}tz]{haresamudram2021contrastive}
Haresamudram, H., Essa, I., and Pl{\"o}tz, T.
\newblock Contrastive predictive coding for human activity recognition.
\newblock \emph{Proceedings of the ACM on Interactive, Mobile, Wearable and Ubiquitous Technologies}, 5\penalty0 (2):\penalty0 1--26, 2021.

\bibitem[He et~al.(2023)He, Queen, Koker, Cuevas, Tsiligkaridis, and Zitnik]{he2023domain}
He, H., Queen, O., Koker, T., Cuevas, C., Tsiligkaridis, T., and Zitnik, M.
\newblock Domain adaptation for time series under feature and label shifts.
\newblock In \emph{International Conference on Machine Learning}, pp.\  12746--12774. PMLR, 2023.

\bibitem[He et~al.(2016)He, Zhang, Ren, and Sun]{he2016deep}
He, K., Zhang, X., Ren, S., and Sun, J.
\newblock Deep residual learning for image recognition.
\newblock In \emph{Proceedings of the IEEE conference on computer vision and pattern recognition}, pp.\  770--778, 2016.

\bibitem[Hendrycks et~al.(2021)Hendrycks, Basart, Mu, Kadavath, Wang, Dorundo, Desai, Zhu, Parajuli, Guo, et~al.]{hendrycks2021many}
Hendrycks, D., Basart, S., Mu, N., Kadavath, S., Wang, F., Dorundo, E., Desai, R., Zhu, T., Parajuli, S., Guo, M., et~al.
\newblock The many faces of robustness: A critical analysis of out-of-distribution generalization.
\newblock In \emph{Proceedings of the IEEE/CVF international conference on computer vision}, pp.\  8340--8349, 2021.

\bibitem[Hinton et~al.(2015)Hinton, Vinyals, and Dean]{hinton2015distilling}
Hinton, G., Vinyals, O., and Dean, J.
\newblock Distilling the knowledge in a neural network.
\newblock \emph{arXiv preprint arXiv:1503.02531}, 2015.

\bibitem[Howard et~al.(2019)Howard, Sandler, Chu, Chen, Chen, Tan, Wang, Zhu, Pang, Vasudevan, et~al.]{howard2019searching}
Howard, A., Sandler, M., Chu, G., Chen, L.-C., Chen, B., Tan, M., Wang, W., Zhu, Y., Pang, R., Vasudevan, V., et~al.
\newblock Searching for mobilenetv3.
\newblock In \emph{Proceedings of the IEEE/CVF international conference on computer vision}, pp.\  1314--1324, 2019.

\bibitem[Jiang et~al.(2023)Jiang, Han, Zhao, and Wang]{jiang2023pdformer}
Jiang, J., Han, C., Zhao, W.~X., and Wang, J.
\newblock Pdformer: Propagation delay-aware dynamic long-range transformer for traffic flow prediction.
\newblock In \emph{Proceedings of the AAAI conference on artificial intelligence}, volume~37, pp.\  4365--4373, 2023.

\bibitem[Karmanov et~al.(2024)Karmanov, Guan, Lu, El~Saddik, and Xing]{karmanov2024efficient}
Karmanov, A., Guan, D., Lu, S., El~Saddik, A., and Xing, E.
\newblock Efficient test-time adaptation of vision-language models.
\newblock In \emph{Proceedings of the IEEE/CVF Conference on Computer Vision and Pattern Recognition}, pp.\  14162--14171, 2024.

\bibitem[Kong et~al.(2023)Kong, Li, Yuan, and Kong]{kong2023convrelu++}
Kong, R., Li, Y., Yuan, Y., and Kong, L.
\newblock Convrelu++: Reference-based lossless acceleration of conv-relu operations on mobile cpu.
\newblock In \emph{Proceedings of the 21st Annual International Conference on Mobile Systems, Applications and Services}, pp.\  503--515, 2023.

\bibitem[Li et~al.(2024{\natexlab{a}})Li, Jiang, Liu, Huang, and Kwok]{li2024leaerning}
Li, T., Jiang, W., Liu, F., Huang, X., and Kwok, J.~T.
\newblock Learning scalable model soup on a single gpu: An efficient subspace training strategy.
\newblock In \emph{European conference on computer vision (ECCV)}, 2024{\natexlab{a}}.

\bibitem[Li et~al.(2024{\natexlab{b}})Li, Li, Li, Cao, and Liu]{li2024flexnn}
Li, X., Li, Y., Li, Y., Cao, T., and Liu, Y.
\newblock Flexnn: Efficient and adaptive dnn inference on memory-constrained edge devices.
\newblock In \emph{Proceedings of the 30th Annual International Conference on Mobile Computing and Networking}, pp.\  709--723, 2024{\natexlab{b}}.

\bibitem[Liu et~al.(2020)Liu, Zhang, and Wang]{liu2020adaptive}
Liu, Y., Zhang, W., and Wang, J.
\newblock Adaptive multi-teacher multi-level knowledge distillation.
\newblock \emph{Neurocomputing}, 415:\penalty0 106--113, 2020.

\bibitem[Lu \& Sun(2024)Lu and Sun]{pmlr-v235-lu24i}
Lu, J. and Sun, S.
\newblock {C}au{D}i{TS}: Causal disentangled domain adaptation of multivariate time series.
\newblock In Salakhutdinov, R., Kolter, Z., Heller, K., Weller, A., Oliver, N., Scarlett, J., and Berkenkamp, F. (eds.), \emph{Proceedings of the 41st International Conference on Machine Learning}, volume 235 of \emph{Proceedings of Machine Learning Research}, pp.\  33113--33146. PMLR, 21--27 Jul 2024.

\bibitem[Lu et~al.(2024)Lu, Zhou, Xu, Zhang, Gao, and Li]{luspp}
Lu, X., Zhou, A., Xu, Y., Zhang, R., Gao, P., and Li, H.
\newblock Spp: Sparsity-preserved parameter-efficient fine-tuning for large language models.
\newblock In Salakhutdinov, R., Kolter, Z., Heller, K., Weller, A., Oliver, N., Scarlett, J., and Berkenkamp, F. (eds.), \emph{Proceedings of the 41st International Conference on Machine Learning}, Proceedings of Machine Learning Research. PMLR, 21--27 Jul 2024.

\bibitem[Ouyang et~al.(2021)Ouyang, Xie, Zhou, Huang, and Xing]{ouyang2021clusterfl}
Ouyang, X., Xie, Z., Zhou, J., Huang, J., and Xing, G.
\newblock Clusterfl: a similarity-aware federated learning system for human activity recognition.
\newblock In \emph{Proceedings of the 19th Annual International Conference on Mobile Systems, Applications, and Services}, pp.\  54--66, 2021.

\bibitem[Ouyang et~al.(2022)Ouyang, Shuai, Zhou, Shi, Xie, Xing, and Huang]{ouyang2022cosmo}
Ouyang, X., Shuai, X., Zhou, J., Shi, I.~W., Xie, Z., Xing, G., and Huang, J.
\newblock Cosmo: contrastive fusion learning with small data for multimodal human activity recognition.
\newblock In \emph{Proceedings of the 28th Annual International Conference on Mobile Computing And Networking}, pp.\  324--337, 2022.

\bibitem[Ouyang et~al.(2023)Ouyang, Xie, Fu, Cheng, Pan, Ling, Xing, Zhou, and Huang]{ouyang2023harmony}
Ouyang, X., Xie, Z., Fu, H., Cheng, S., Pan, L., Ling, N., Xing, G., Zhou, J., and Huang, J.
\newblock Harmony: Heterogeneous multi-modal federated learning through disentangled model training.
\newblock In \emph{Proceedings of the 21st Annual International Conference on Mobile Systems, Applications and Services}, pp.\  530--543, 2023.

\bibitem[Pan \& Yang(2009)Pan and Yang]{pan2009survey}
Pan, S.~J. and Yang, Q.
\newblock A survey on transfer learning.
\newblock \emph{IEEE Transactions on knowledge and data engineering}, 22\penalty0 (10):\penalty0 1345--1359, 2009.

\bibitem[Peng et~al.(2024)Peng, Fang, Zhang, and Lu]{pmlr-v235-peng24a}
Peng, B., Fang, Z., Zhang, G., and Lu, J.
\newblock Knowledge distillation with auxiliary variable.
\newblock In Salakhutdinov, R., Kolter, Z., Heller, K., Weller, A., Oliver, N., Scarlett, J., and Berkenkamp, F. (eds.), \emph{Proceedings of the 41st International Conference on Machine Learning}, volume 235 of \emph{Proceedings of Machine Learning Research}, pp.\  40185--40199. PMLR, 21--27 Jul 2024.

\bibitem[Phan et~al.(2024)Phan, Brantley, Milani, Mehri, Swamy, and Gordon]{pmlr-v235-phan24a}
Phan, M., Brantley, K., Milani, S., Mehri, S., Swamy, G., and Gordon, G.~J.
\newblock When is transfer learning possible?
\newblock In Salakhutdinov, R., Kolter, Z., Heller, K., Weller, A., Oliver, N., Scarlett, J., and Berkenkamp, F. (eds.), \emph{Proceedings of the 41st International Conference on Machine Learning}, volume 235 of \emph{Proceedings of Machine Learning Research}, pp.\  40642--40666. PMLR, 21--27 Jul 2024.

\bibitem[Pizzolato et~al.(2017)Pizzolato, Tagliapietra, Cognolato, Reggiani, M{\"u}ller, and Atzori]{pizzolato2017comparison}
Pizzolato, S., Tagliapietra, L., Cognolato, M., Reggiani, M., M{\"u}ller, H., and Atzori, M.
\newblock Comparison of six electromyography acquisition setups on hand movement classification tasks.
\newblock \emph{PloS one}, 12\penalty0 (10):\penalty0 e0186132, 2017.

\bibitem[Qu et~al.(2024)Qu, Zou, He, R\"ohrbein, Knoll, Chen, and Jiang]{Qu2024LEAD}
Qu, S., Zou, T., He, L., R\"ohrbein, F., Knoll, A., Chen, G., and Jiang, C.
\newblock Lead: Learning decomposition for source-free universal domain adaptation.
\newblock In \emph{Proceedings of the IEEE/CVF Conference on Computer Vision and Pattern Recognition (CVPR)}, pp.\  23334--23343, June 2024.

\bibitem[Saeed et~al.(2019)Saeed, Ozcelebi, and Lukkien]{saeed2019multi}
Saeed, A., Ozcelebi, T., and Lukkien, J.
\newblock Multi-task self-supervised learning for human activity detection.
\newblock \emph{Proceedings of the ACM on Interactive, Mobile, Wearable and Ubiquitous Technologies}, 3\penalty0 (2):\penalty0 1--30, 2019.

\bibitem[Szegedy et~al.(2015)Szegedy, Liu, Jia, Sermanet, Reed, Anguelov, Erhan, Vanhoucke, and Rabinovich]{szegedy2015going}
Szegedy, C., Liu, W., Jia, Y., Sermanet, P., Reed, S., Anguelov, D., Erhan, D., Vanhoucke, V., and Rabinovich, A.
\newblock Going deeper with convolutions.
\newblock In \emph{Proceedings of the IEEE conference on computer vision and pattern recognition}, pp.\  1--9, 2015.

\bibitem[Tan et~al.(2018)Tan, Sun, Kong, Zhang, Yang, and Liu]{tan2018survey}
Tan, C., Sun, F., Kong, T., Zhang, W., Yang, C., and Liu, C.
\newblock A survey on deep transfer learning.
\newblock In \emph{Artificial Neural Networks and Machine Learning--ICANN 2018: 27th International Conference on Artificial Neural Networks, Rhodes, Greece, October 4-7, 2018, Proceedings, Part III 27}, pp.\  270--279. Springer, 2018.

\bibitem[Tong et~al.(2021)Tong, Xu, Huang, and Zheng]{tong2021mathematical}
Tong, X., Xu, X., Huang, S.-L., and Zheng, L.
\newblock A mathematical framework for quantifying transferability in multi-source transfer learning.
\newblock \emph{Advances in Neural Information Processing Systems}, 34:\penalty0 26103--26116, 2021.

\bibitem[Vaswani(2017)]{vaswani2017attention}
Vaswani, A.
\newblock Attention is all you need.
\newblock \emph{Advances in Neural Information Processing Systems}, 2017.

\bibitem[Vemulapalli et~al.(2024)Vemulapalli, Pouransari, Faghri, Mehta, Farajtabar, Rastegari, and Tuzel]{knowledgetransfer2024}
Vemulapalli, R., Pouransari, H., Faghri, F., Mehta, S., Farajtabar, M., Rastegari, M., and Tuzel, O.
\newblock Knowledge transfer from vision foundation models for efficient training of small task-specific models.
\newblock In \emph{International Conference on Machine Learning (ICML)}, 2024.

\bibitem[Wen et~al.(2023)Wen, Li, Zhang, Jiang, Ye, Ouyang, Zhang, and Liu]{wen2023adaptivenet}
Wen, H., Li, Y., Zhang, Z., Jiang, S., Ye, X., Ouyang, Y., Zhang, Y., and Liu, Y.
\newblock Adaptivenet: Post-deployment neural architecture adaptation for diverse edge environments.
\newblock In \emph{Proceedings of the 29th Annual International Conference on Mobile Computing and Networking}, pp.\  1--17, 2023.

\bibitem[Wilson et~al.(2021)Wilson, Doppa, and Cook]{wilson2021calda}
Wilson, G., Doppa, J.~R., and Cook, D.~J.
\newblock Calda: Improving multi-source time series domain adaptation with contrastive adversarial learning.
\newblock \emph{arXiv preprint arXiv:2109.14778}, 2021.

\bibitem[Xu et~al.(2021)Xu, Zhou, Tan, Li, and Shen]{xu2021limu}
Xu, H., Zhou, P., Tan, R., Li, M., and Shen, G.
\newblock Limu-bert: Unleashing the potential of unlabeled data for imu sensing applications.
\newblock In \emph{Proceedings of the 19th ACM Conference on Embedded Networked Sensor Systems}, pp.\  220--233, 2021.

\bibitem[Zhang et~al.(2022)Zhang, Chen, and Wang]{zhang2022confidence}
Zhang, H., Chen, D., and Wang, C.
\newblock Confidence-aware multi-teacher knowledge distillation.
\newblock In \emph{ICASSP 2022-2022 IEEE International Conference on Acoustics, Speech and Signal Processing (ICASSP)}, pp.\  4498--4502. IEEE, 2022.

\bibitem[Zhu et~al.(2018)Zhu, Liu, Liu, Hu, Wang, Tan, Huang, Zhu, Ji, Jiang, et~al.]{zhu2018deep}
Zhu, N., Liu, X., Liu, Z., Hu, K., Wang, Y., Tan, J., Huang, M., Zhu, Q., Ji, X., Jiang, Y., et~al.
\newblock Deep learning for smart agriculture: Concepts, tools, applications, and opportunities.
\newblock \emph{International Journal of Agricultural and Biological Engineering}, 11\penalty0 (4):\penalty0 32--44, 2018.

\bibitem[Zhu et~al.(2020)Zhu, Zhuang, Wang, Ke, Chen, Bian, Xiong, and He]{zhu2020deep}
Zhu, Y., Zhuang, F., Wang, J., Ke, G., Chen, J., Bian, J., Xiong, H., and He, Q.
\newblock Deep subdomain adaptation network for image classification.
\newblock \emph{IEEE transactions on neural networks and learning systems}, 32\penalty0 (4):\penalty0 1713--1722, 2020.

\end{thebibliography}
\bibliographystyle{icml2025}

\newpage
\appendix
\onecolumn
\section{Details of Experiment Settings}
\subsection{Datasets.}
\label{appendix: datasets}

HaT is evaluated on the four datasets in Table \ref{tab: dataset}, which are further introduced as follows:

\textbf{HARBox \cite{ouyang2021clusterfl}.} This dataset consists of 9-axis Inertial Measurement Unit (IMU) data collected via crowdsourcing from 120 users. It includes data for five activities, such as walking and hopping.

\textbf{ImageNet-R \cite{hendrycks2021many}.} This dataset contains over 30k images from 200 classes in 16 different styles. Each style can be considered a small dataset. We filtered out styles with limited data or unclear labels, resulting in 8 styles for experiments.

\textbf{NinaPro \cite{pizzolato2017comparison}.}  This dataset contains the electromyogram (EMG) data collected from 10 subjects. Two commercial EMG sensors, the Myo Armbands, are deployed around the elbows of the subjects for 6-class gesture recognition.

\textbf{Alzheimer's Disease (AD) \cite{ouyang2023harmony}.} This dataset consists of Alzheimer's Disease-related activity data collected from 16 home environments using multiple modalities. It includes 11 activity classes, such as writing and sleeping.

While HaT is evaluated on these three diverse applications, it has the potential to extend to other learning systems, such as traffic management or smart agriculture \cite{jiang2023pdformer, zhu2018deep}, which we plan to explore in future work.

\begin{table*}[!ht]
\centering
\caption{Datasets for evaluating the effectiveness and generalizability of HaT.}
\scalebox{0.85}{\begin{tabular}{cccccc}
\toprule
Dataset    & Class & Domain (\#) & Task                       & Modality  & Model Library\\ \midrule
HARBox     & 5     & 120 users      & Human Activity Recognition & IMU    & TPN-(S, M, L), CPC-(S, M, L)              \\
ImageNet-R & 200   & 8 styles & Image Classification       & Image & GoogleNet, MobileNet(S, L), ResNet-(18, 34, 50)\\
NinaPro    & 6     & 10 users & Gesture Recognition        & EMG & ConvNet-(S, M, L), RNN-(S, M, L)\\
\multirow{2}{*}{AD}         & \multirow{2}{*}{11}    & \multirow{2}{*}{16 users} & \multirow{2}{*}{Human Activity Recognition} & Depth Camera, & \multirow{2}{*}{ADNet-(S, M, L), TinyADNet-(S, M, L)} \\
 &    &  &  & Audio, Radar &  \\
\bottomrule
\end{tabular}\label{tab: dataset}}
\end{table*}

\subsection{Model Libraries.} \label{app: model library}
We include six different models for each dataset. For IMU data, the TPN-(S, M, L) \cite{saeed2019multi} and CPC-(S, M, L) \cite{haresamudram2021contrastive} models are used, with feature channels of 12, 16, 32 for TPN and 8, 12, 16 for CPC, respectively. For image processing, the model library consists of GoogleNet \cite{szegedy2015going}, MobileNet-v3 (S, L) \cite{howard2019searching}, and ResNet-(18, 34, 50) \cite{he2016deep}. For EMG data, the ConvNet-(S, M, L) \cite{cote2019deep} models and a RNN-(S, M, L) are utilized, with feature channels of 4, 8, 12 for ConvNet and 32, 48, 56 for RNN, respectively. To handle the multi-modal data in AD, we adapt the model in \cite{ouyang2023harmony} by varying the number of layers and feature dimensions, creating 5-layer ADNet-(S, M, L) and 3-layer TinyADNet-(S, M, L) models with 64, 128, 256 for ADNet and 32, 64, 96 feature channels for TinyADNet.

\subsection{Baselines}
\label{appendix: baselines}
HaT is compared with five most relevant baselines covering the area of knowledge distillation, domain adapation, and model aggregation:

\textbf{DistillWeighted \cite{borup2023distilling}.} DistillWeighted uses existing vision models to build models for new tasks. Based on the PARC metric \cite{bolya2021scalable}, it assigns fixed weights to combine the predictions of all source models for knowledge distillation. As executing all source models is too expensive, we pre-select $N_p$ models using the PARC metric and then apply DistillWeighted.

\textbf{DistillNearest \cite{borup2023distilling}.} DistillNearest selects a single model from the most similar source domain based on the PARC metric. The target model then learns from the pseudo labels generated by the selected model and the labeled data.

\textbf{LEAD \cite{Qu2024LEAD}.} LEAD is a domain adaptation method that adapts the source model to builds instance-level decision boundary for target data using decomposed source features.

\textbf{MEHLSoup \cite{li2024leaerning}.} MEHLSoup merges multiple source domain models with a learned mixing coefficient, which is optimized by a block coordinate gradient descent algorithm on the target domain data.

\textbf{AccDistill.} We select and ensemble source domain models with top-k accuracy and then transfer the knowledge from the ensembled model to the target models, leveraging the distilltion methods in \cite{borup2023distilling}.

Other knowledge distillation, domain adaptation, or model merging methods are not included, as they have already been outperformed by the considered baselines \cite{borup2023distilling, Qu2024LEAD, li2024leaerning}. Since LEAD and MEHLSoup, as well as other adaptation and merging methods, are unable to handle model heterogeneity, they are not directly comparable to HaT. To make both methods executable, we select source domains with architectures that match the target models as candidates. Federated learning methods are not included for comparison due to the difference in the considered scenario (See Section \ref{sec: problem formulation}). Self-supervised learning methods \cite{xu2021limu, ouyang2022cosmo} are not included as baselines as they are orthogonal to HaT and can be combined to further enhance performance. 


\subsection{Data Splits and Training Details in MRSE}
\label{app: data splits MRSE}
The domains in each dataset are randomly divided into five groups, denoted as $\{G(i), i=0, \cdots, 4\}$. Detailed information about the groups is provided in Table \ref{tab: group infor}. In round $j$, the domains in $\{G(i), i=0, \cdots, j-1\}$ serve as the source domains $\mathcal{D}^S(j)$, while the domains in $G(j)$ are the target domains $\mathcal{D}^T(j)$. Once the models in $G(j)$ are ready for use, they are incorporated as source domains in the subsequent round $j+1$, sharing their knowledge with new targets for further expansion. For instance, during round 2 expansion on the HARBox dataset, the source domains include 60 users from $G(0)$ and $G(1)$, whose knowledge is used to build models for 20 users in $G(2)$. After round 2 expansion, the system scales from 60 to 80 users, and the models in $G(2)$ are subsequently leveraged to construct models for $G(3)$ along with $G(0)$ and $G(1)$. The model skeletons for all domains are randomly selected from TPN-(S, M, L), ResNet-(18, 34, 50), ConvNet-(S, M, L), and ADNet-(S, M, L) for the four datasets, respectively.

The source domain models are trained using supervised learning on the labeled data of each domain. For the target domains, 60\% of the data is randomly selected as the training set, 20\% as the validation set, and the remaining 20\% as the test set. The parameter $\gamma$ is set to 10\%.

\begin{table}[!tb]
\centering
\caption{Group information for the four datasets.} \label{tab: group infor}
\begin{tabular}{c|cccc}
\toprule
                          & HARBox & ImageNet-R & NinaPro & AD  \\\midrule
Groups & [40, 20, 20, 20, 20]  & [4, 1, 1, 1, 1]    & [6, 1, 1, 1, 1] & [8, 2, 2, 2, 2] \\\bottomrule
\end{tabular}
\end{table}

\begin{figure}[!tb]
    \centering
    \includegraphics[width=0.6\linewidth]{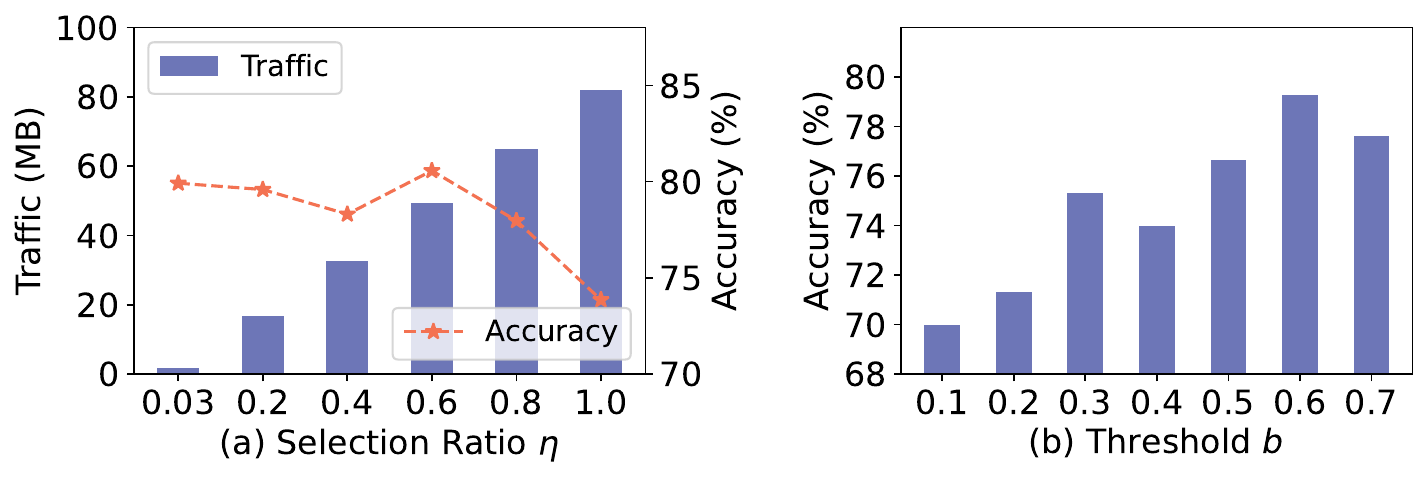}
    \caption{Sensitivity Analysis. (a) The impact of the selection ratio $\eta$. (b) The impact of the threshold $b$.}
    \label{fig: selection ratio}
\end{figure}

\subsection{Data Splits and Training Details in OTSE}
\label{app: data splits OTSE}
We also compare the results of HaT against the baselines in the OTSE setting, where one domain is randomly selected as the target domain, and the remaining domains serve as source domains. The source domain architectures are randomly selected from TPN-(S, M, L), ResNet-(18, 34, 50), ConvNet-(S, M, L), and ADNet-(S, M, L). The other settings are kept aligned with those in Section \ref{app: data splits MRSE}. For ImageNet-R, each of the eight styles is tested separately. For the other three datasets, ten different splits are randomly generated, and the average accuracy and communication overhead are reported. 

\section{Sensitivity Analysis}
\label{app: Sensitivity Analysis}
We further conduct a sensitivity analysis on two key parameters in HaT, the selection ratio $\eta$ and the threshold $b$.

\textbf{Varied Selection Ratio.} The impact of the Feature-based Coarse Selection (FbCS) on communication overhead is illustrated in Figure \ref{fig: selection ratio}(a) by varying its selection ratio, $\eta$. Figure \ref{fig: selection ratio}(a) shows that as $\eta$ decreases from 1.0 (without FbCS) to 0.03 (without Centroids-Accuracy Joint Selection), the communication traffic consistently decreases because fewer models are selected for transmission. The model execution cost also decreases as fewer source models are chosen for encoding data in the target domain. Additionally, Figure \ref{fig: selection ratio}(a) shows that the accuracy achieved by the target model increases and then slightly decreases as $\eta$ decreases. This pattern occurs because, when $\eta$ is large, the FbCS filters out less useful models. However, when $\eta$ becomes too small, the coarse selection inadvertently discard some high-quality models.

\textbf{Varied Threshold.} Figure \ref{fig: selection ratio}(b) illustrates that as the threshold $b$ increases, the average accuracy of the target models initially improves but eventually declines. A small threshold results in frequent model updates early in training, during which the aggregated predictions from the mixer are of low quality, leading to suboptimal performance. Conversely, an excessively large threshold causes the adaptive knowledge injection process to degrade into direct training with limited labeled data, thereby failing to leverage the knowledge from source models. Since the quality of selected models varies across domains and tasks, we recommend setting the threshold slightly higher than the highest accuracy of the selected models on the labeled target data. This recommendation is based on the insight that knowledge from different models can complement one another to improve overall performance.



\begin{table*}[!t]
\caption{
Detailed performance comparison on the HARBox dataset in the MRSE setting.}
    \label{tab: system expansion on HARBox}
    \scalebox{0.93}{
    \begin{tabular}{c|cc|cc|cc|cc|ccc}
    \toprule
    \multirow{2}{*}{Methods} & \multicolumn{2}{c|}{Round 1}   & \multicolumn{2}{c|}{Round 2}      & \multicolumn{2}{c|}{Round 3}   & \multicolumn{2}{c|}{Round 4}  & Average & Total \\ 
    &  Accuracy            & Traffic             & Accuracy            & Traffic             & Accuracy            & Traffic             & Accuracy            & Traffic & Accuracy & Traffic \\\midrule
     LEAD & 31.42 & 208 & 43.60 & 388 & 43.35 & 532 & 40.30 & 355  & 39.67 & 1483 \\ 
     MEHLSoup & 57.63 &  208 & 65.84 & 388 & 63.51 & 532 & 64.95 & 355 & 62.98 & 1483 \\ \midrule
     AccDistill & 68.71 &  490 & 76.36 & 830 & 75.34 & 1188 & 73.20 & 1526 & 73.40 & 4034 \\  
     DistillNearest & 73.57 & 490 & 77.87 & 830 & 76.36 & 1188 & 72.04 & 1526  & 74.95 & 4034 \\ 
     DistillWeighted & 72.91 &  490 & 77.67 & 830 & 76.54 & 1188 & 74.54 & 1526 & 75.42 & 4034\\ \midrule
     HaT & \textbf{77.74} & \textbf{296} & \textbf{81.05} & \textbf{534} & \textbf{79.42} & \textbf{741} & \textbf{78.85} & \textbf{946} & \textbf{79.27} & \textbf{2517} \\
    \bottomrule
    \end{tabular}
    }
\end{table*}

\begin{figure*}[!tb]
        \centering

        \subfigure[Varied target models.]{
            \centering
            \includegraphics[width=0.45\linewidth]{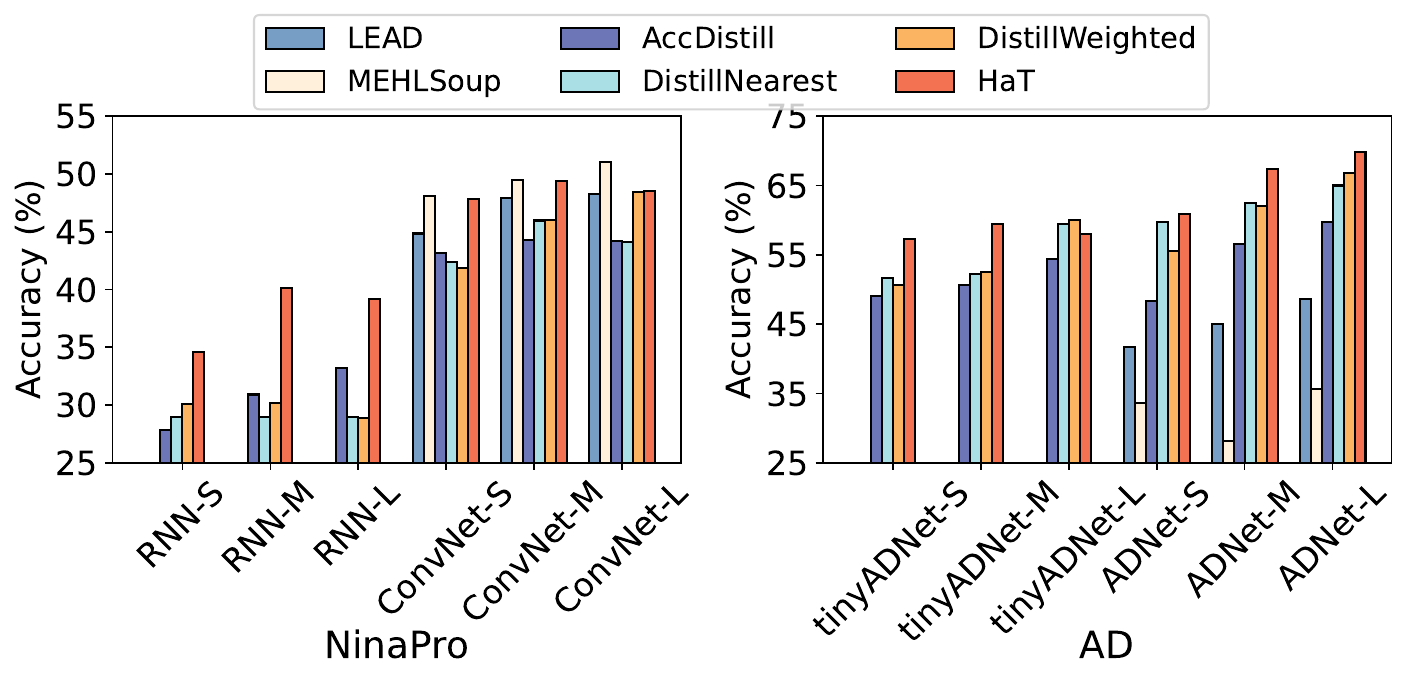}
            \label{fig: archi app}
            }
        \hspace{0mm}
        \subfigure[Varied portion $\gamma$.]{
            \centering
            \includegraphics[width=0.45\linewidth]{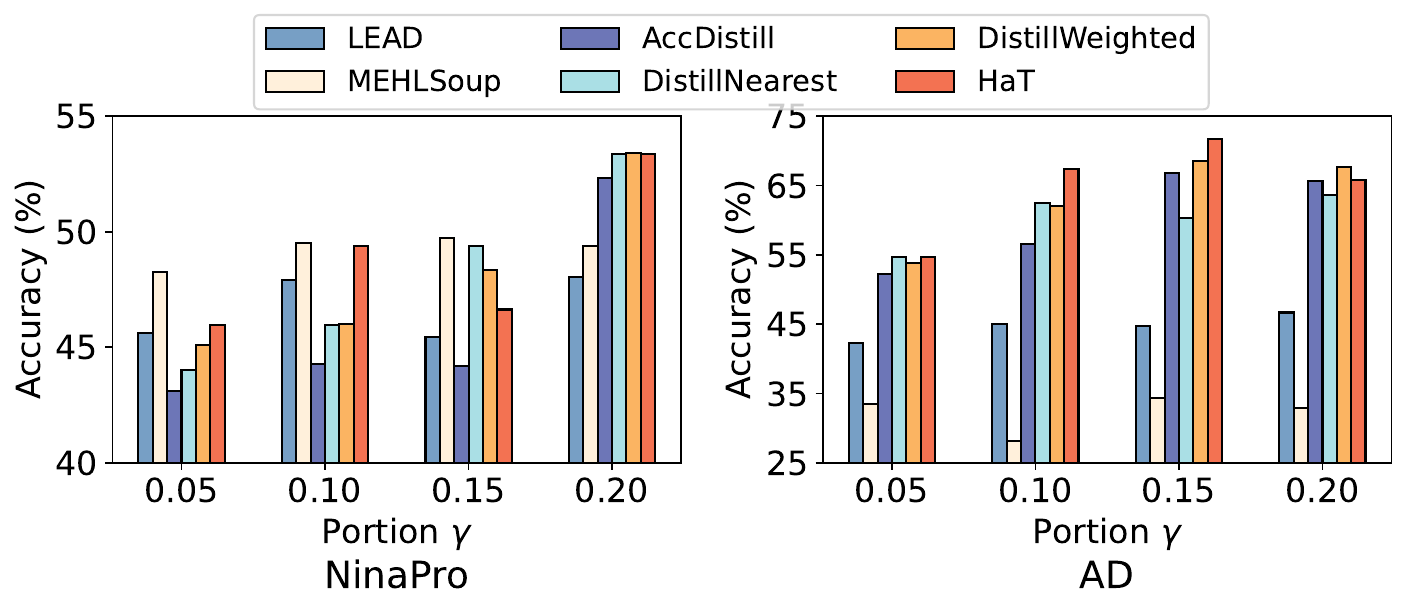}
            \label{fig: portion app}
            }
            
    \caption{Results Comparsion on NinaPro and AD in the OTSE setting. (a) The target model skeleton are varied. (b) The portion of labeled data in the target domain are varied.} 
\end{figure*}

\section{Detailed Results Comparison in the MRSE setting}
\label{app: multi-round results MRSE}
Table \ref{tab: system expansion on HARBox} presents detailed results on each rounds of expansion on HARBox in the MRSE setting. Similar results are observed on the other datasets. HaT outperforms the best baselines by 4.2\%, 3.2\%, 3.1\%, and 4.3\% in accuracy from round 1 to 4, respectively. Besides, the traffic during expansion is significant less compared with the methods that does not have constraints in the source model architectures (LEAD and MEHLSoup only leverage source models that share the same architectures with the target models, thus their traffic are not directly comparable with HaT).

\section{Additional Results in the OTSE setting}
\label{app: additional results OTSE}
Addtional results on NinaPro and AD are presented in Figure \ref{fig: archi app} and Figure \ref{fig: portion app} when the target architectures and the portion of labeled data $\gamma$ are varied. 

As shown in Figure \ref{fig: archi app} and Figure \ref{fig: portion app}, HaT achieves superior or comparable performance in most cases. Although HaT performs slightly worse than MEHLSoup on NinaPro, it significantly outperforms MEHLSoup on the other datasets, likely due to the lower data heterogeneity in NinaPro, which aligns better with MEHLSoup's approach.

\end{document}